%% file: root.tex
\documentclass[letterpaper, 10 pt, conference]{ieeeconf}  % Comment this line out if you need a4paper

\IEEEoverridecommandlockouts                              % This command is only needed if 
\usepackage{graphics} % for pdf, bitmapped graphics files
\usepackage{epsfig} % for postscript graphics files
\usepackage{mathptmx} % assumes new font selection scheme installed
\usepackage{times} % assumes new font selection scheme installed
\usepackage{amsmath} % assumes amsmath package installed
\usepackage{amssymb}  % assumes amsmath package installed

\let\labelindent\relax % Fix for enumitem conflict with ieeeconf
\usepackage[inline]{enumitem}
\usepackage{array}
\usepackage{booktabs}
\title{\LARGE \bf
ACTLLM: Action Consistency Tuned Large Language Model\thanks{This work is a work in progress.}}

\author{Jing Bi$^{1}$, Lianggong Bruce Wen$^{2}$, Zhang Liu$^{2}$, Chenliang Xu$^{1}$% <-this % stops a space
\thanks{$^{1}$University of Rochester, Rochester, NY, USA. Email: jing.bi@rochester.edu, chenliang.xu@rochester.edu}%
\thanks{$^{2}$Corning Inc., Corning, NY, USA. Email: \{wenl, liuzh3\}@corning.com}%
}

\begin{document}

\maketitle
\thispagestyle{empty}
\pagestyle{empty}
\input{sec/0_abstract}
\input{sec/1_intro}
\input{sec/2_related}
\input{sec/3_method}

\input{sec/4_exp}
\input{sec/5_conclusion}
\newpage

\bibliographystyle{IEEEtran}
\bibliography{aaai25}  % Use your .bib files as needed

% \addbibresource{aaai25.bib}
\end{document}

%% file: sec/0_abstract.tex
\begin{abstract}
This paper introduces ACTLLM (Action Consistency Tuned Large Language Model), a novel approach for robot manipulation in dynamic environments.
Traditional vision-based systems often struggle to learn visual representations that excel in both task execution and spatial reasoning, thereby limiting their adaptability in dynamic environments. 
ACTLLM addresses these challenges by harnessing language to craft structured scene descriptors, providing a uniform interface for both spatial understanding and task performance through flexible language instructions. 
Moreover, we introduce a novel action consistency constraint that aligns visual perception with corresponding actions, thereby enhancing the learning of actionable visual representations. 
Additionally, we have reformulated the Markov decision process for manipulation tasks into a multi-turn visual dialogue framework. 
This approach enables the modeling of long-term task execution with enhanced contextual relevance derived from the history of task execution. 
During our evaluation, ACTLLM excels in diverse scenarios, proving its effectiveness on challenging vision-based robot manipulation tasks.

% integrating spatial reasoning with visual-language instructions, 
% We first 

% then we employs a 

% Key contributions include: 1) Combining spatial reasoning with language for contextual relevance and practical applicability, 2) Addressing the challenge of state representation learning through language model tuning, and 3) Enhancing LLM fine-tuning with manipulation control data. 

\end{abstract}

%% file: sec/1_intro.tex
\section{Introduction}
% problem statement, two problems
Adapting to diverse and dynamic environments, while executing flexible task specifications, remains a significant challenge in robot manipulation methods. 
Language-based vision manipulation systems offer a promising solution by leveraging rich visual information and language to better comprehend varying environmental conditions and task specifications~\cite{jiang2022vima,cliport,li2023mastering,mees2022calvin}.
These systems typically comprise two key components: a vision component that associates concepts from language instructions with visual information, and a policy module that generates actions based on the outputs from the vision component.
Recent end-to-end frameworks~\cite{cliport,Ahn2022,driess2023palm,brohan2022rt,tellex2011understanding} attempt to leverage affordances to integrate semantic meanings with visual information.
This integration helps robots identify feasible actions within specific physical contexts, addressing the question of which actions are possible and where they can be performed in a given scene by merging linguistic context with visual cues.
In contrast, another line of research, exemplified by CALVIN\cite{mees2022calvin} advocates for a transition towards adaptable, task-agnostic manipulation policies that employ general, unstructured language to define tasks. This approach allows for more generalized policies as the broad language specifications enable versatility in task execution.
\begin{figure}[t!]
	\centering
	\includegraphics[width=0.9\columnwidth]{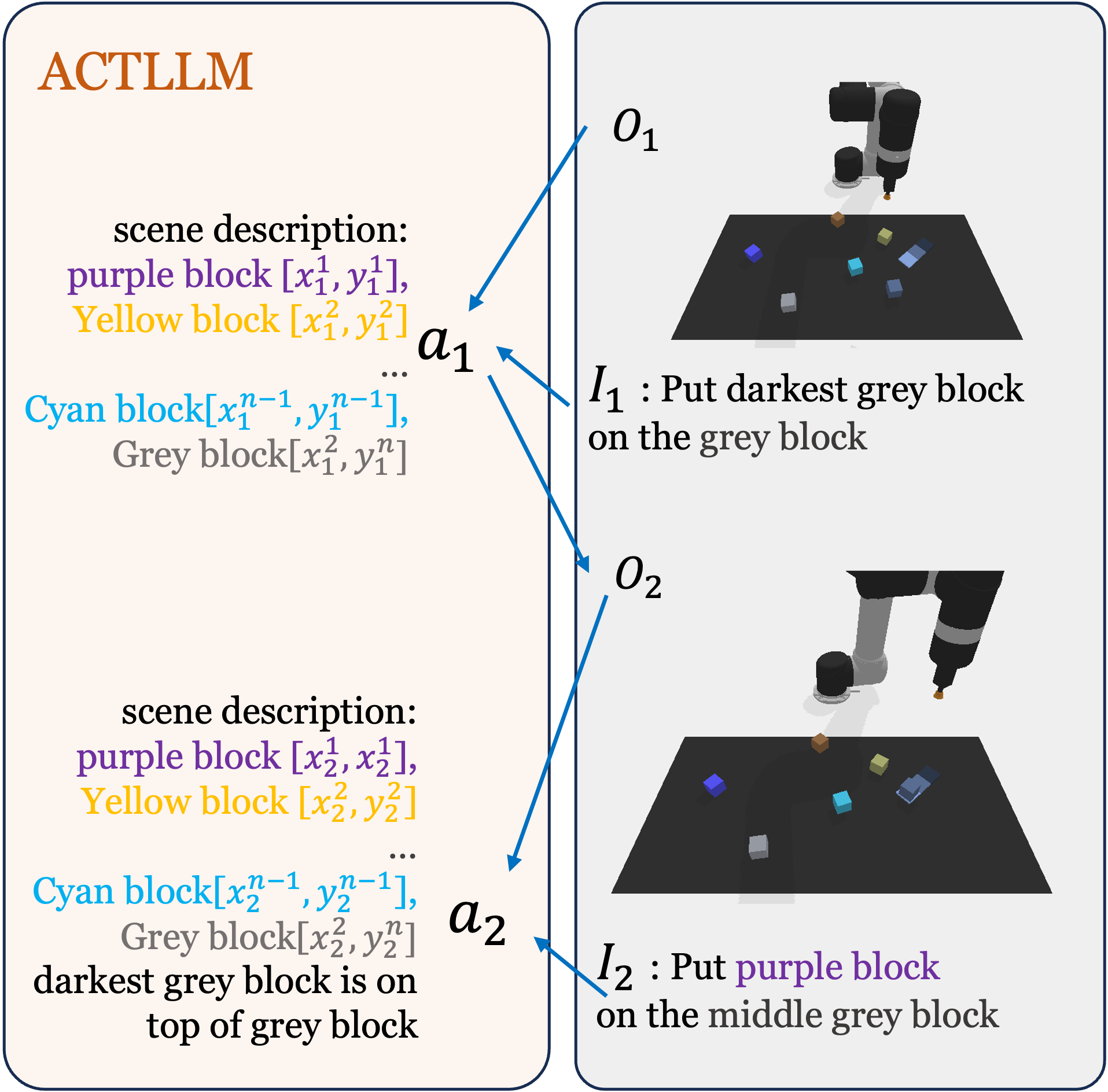}
	\caption{We reformulated the Markov decision process for manipulation tasks into a multi-turn visual dialogue framework, wherein the model generates descriptions of the current view and potential future states, based on given instructions and history observations. From these descriptions, actions are generated to accomplish the tasks. This approach ensures that the actions generated between scene descriptions are consistent with the changes in the scene, empowering Large Language Models (LLMs) to analyze spatial and temporal relationships within the manipulation trajectory. This facilitates the efficient execution of complex tasks.}
	\label{fig:conv}
    \vspace{-5mm}
\end{figure}

However, existing research frequently separates the learning processes of vision models and action policies, with each component being optimized independently.
This division often results in perception models often provides representations that are not optimized to be effectively used by the policy models.
Moreover, the action policies face the challenge of learning a broad spectrum of skills.
This complexity arises from the necessity to interpret open-ended language instructions and effectively leverage visual representations, thus demanding more sophisticated and integrated learning approaches.
Therefore, achieving more adaptable policies that associate visual observations with linguistic concepts for robot manipulation presents a two-stage challenge:
\begin{enumerate*}[label=(\roman*)]
    \item Enhancing the alignment between the vision and language concepts.
    \item Optimizing the integrated information from both modalities for action policy.
\end{enumerate*}

% Recent advancements in foundational models~\cite{chiang2023vicuna,clip,slip} have significantly contributed to overcoming the first challenge, particularly in enhancing robots' abilities to comprehend tasks, follow user instructions, and interpret their surrounding environments
Due to the strong zero-shot performance of foundation models, several works have integrated these models into robot applications to overcome the first challenge.
These methods encompass various tasks ranging from generating natural language scene descriptions~\cite{huang2023instruct2act}, leveraging pretrained representation for reactive control~\cite{nair2022r3m}, grounding actions to symbolic representation for test-time adaptation\cite{ge2023policy,wang2023programmatically,zhang2023sprint} to automating dense reward generation~\cite{ma2023liv,Kwon2023,Yu2023,Biyik2022,ma2023liv}.
Moreover, several methods have adapted foundational model architectures to address the second challenge. 
For instance, the ReAct~\cite{yao2022react} planner generates conditional sequences that guide low-level policy actions. 
PaLM-E ~\cite{driess2023palm} integrates multimodal inputs—including vision, text, and state estimation—to produce low-level instructional texts to drive the controller. 
However, previous methods often treat task specification and environment understanding as distinct challenges, a separation that can hinder the integration of visual concepts with their semantic meanings, thereby creating barriers to effectively instructing robots.
Moreover, visual observations are frequently processed into a hidden representation, which is not only difficult to interpret but also challenging to be optimized for policy.

To overcome these challenges, we introduce ACTLLM, a method that unifies the interpretation of visual information with policy learning. 
Our approach leverages structured scene descriptions that are not only more accessible for humans but also allow us to construct a novel loss functions to optimize the model more effectively. 
By incorporating an action consistency constraint loss, we jointly optimize the action policy with scene description generation, significantly enhancing the fusion of these elements and addressing the aforementioned issues.
Furthermore, following proposed method, we are able to reformulated the Markov decision process for manipulation tasks into a multi-turn visual dialogue framework to help long-term task learning.
To summarize, our contributions lay in three folds:

\noindent \textbf{Structured scene description}: By explicitly representing observations as structured scene descriptions, we can integrate spatial reasoning with instruction understanding, creating actionable representations that merge language comprehension with visual features.

\noindent \textbf{Noval Approach to policy optimization:} Building on the foundation of structured scene descriptions, we propose a novel loss function. This enhances the integration of information from both textual instructions and visual observations, optimizing policy learning in dynamic environments.

\noindent \textbf{Enhancing LLM tuning for manipulation} ACTLLM transforms traditional Markov Decision Processes into a visual multi-turn dialogue framework. This enables a novel method for enhancing LLM fine-tuning, incorporating manipulation control data to refine decision-making processes.

%% file: sec/2_related.tex
\begin{figure}[t!]
	\centering
	\includegraphics[width=1\linewidth]{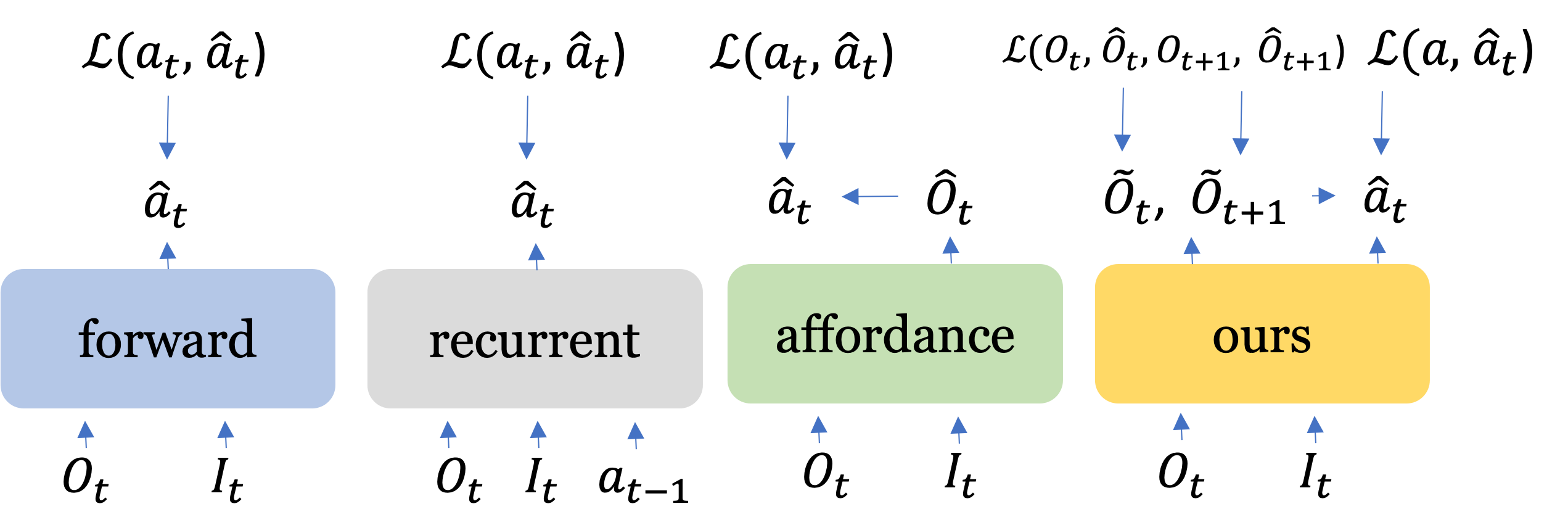}
	\caption{In comparison to previous models: (a) Traditional forward models use current observations and instructions to output actions, with potential enhancements from additional historical information. (b) Affordance methods differ by translating observations into heatmaps for simpler action computation via argmax. (c) Our approach advances these concepts by generating actions from LLMs, utilizing text-based scene descriptions to regularize and improve the accuracy of the action embeddings.}
	\label{fig:relabel}

    \vspace{-5mm}
\end{figure}
\section{Related work}
% \begin{figure*}
% 	\centering
% 	\includegraphics[width=0.9\linewidth]{pic/example.png}
% 	\caption{}
% 	\label{fig:example}
% \end{figure*}

\subsection{Vision for Manipulation} 
Traditional methods for robot perception have primarily relied on explicit 'object' representations, such as instance segmentation, object classes, and poses ~\cite{deng2020self, xie2020best}. However, these methods encounter challenges when dealing with deformable and granular items like clothing and beans, which are difficult to characterize using geometric models or segmentations. To overcome these limitations, recent approaches ~\cite{kalashnikov2018qt, kang2019learning} have begun to make fewer assumptions about objects and tasks, often framing the problem as an image-to-action prediction task. Nevertheless, direct training on RGB images for tasks with 6 Degrees of Freedom (6-DoF) tends to be inefficient, typically necessitating numerous demonstrations or episodes to acquire basic skills like object rearrangement. In response to these challenges, several methods have emerged. For example, in 3D environments, C2FARM ~\cite{james2022coarse} presents an action-centric reinforcement learning (RL) agent with a coarse-to-fine-grain 3D-UNet backbone. However, this approach has a limited receptive field at the finest level, preventing it from encompassing the entire scene. Another line of research, exemplified by approaches like ~\cite{zeng2021transporter, cliport, stengel2022guiding}, focuses on learning action-centric representations with affordances, emphasizing the interaction aspects of objects. In contrast to these methods, our model relies solely on RGB images without the need for special modeling of "objects," as in previous works.

\subsection{Language empowered Robotics} 
Instruction-based policies have been a popular area of research in robotics~\cite{brohan2023can, shao2021concept2robot, shridhar2018interactive}. 
A notable development is CLIPORT~\cite{shridhar2022cliport}, which enhances Transporter~\cite{zeng2021transporter} with semantic understanding and object manipulation abilities via CLIP text encoding~\cite{radford2021learning}. 
This concept was further expanded to 3D environments in the Perceiver-Actor model, utilizing voxelized observation and 3D action spaces~\cite{shridhar2023perceiver}. Another innovative application of LLMs in robotics involves code generation for action policy\cite{liang2022code}, enabling robots to utilize vision APIs for tasks like segmentation, detection, and internet access. 
Instruct2Act~\cite{huang2023instruct2act} exemplifies this by integrating robotic skills with LLMs, striking a balance between flexibility and expertise, and demonstrating high performance in zero-shot settings. 
Language also plays a crucial role in high-level robotic planning~\cite{huang2022language, huang2022inner, ahn2022can} and low-level policy development, with model-based approaches gaining traction~\cite{nair2022learning}. 
Notably, CaP~\cite{liang2022code} and Socratic Models~\cite{zeng2022socratic} have made significant strides by generating detailed policy codes and incorporating perceptual data into LLMs, respectively. Additionally, LLMs are being utilized as a source of reward or feedback in robotic systems, with methods like those by~\cite{huang2022inner} and~\cite{kwon2023reward} enhancing robotic operations through closed-loop feedback and reinforcement learning. 
PAFF ~\cite{ge2023policy} tackles the challenge by utilizing feedback from foundation models via a Hindsight Experience Replay process, enabling the model to respond to randomly generated instructions in unfamiliar environments, with actions subsequently collected and relabeled to fine-tune the foundation model for test-time adaptation. VIMA~\cite{jiang2022vima} proposes the VisuoMotor Attention agent to solve robot manipulation from multimodal prompts with a Transformer Encoder-Decoder architecture. 
% It encodes the multimodal prompts that interleave textual and visual tokens with a pretrained LM by following the practice of Frozen~\cite{tsimpoukelli2021multimodal}.
% In practical applications, RT-1~\cite{brohan2022rt}has effectively showcased the utility of this approach in robotic control which efficiently tokenizes robot inputs and output actions, facilitating real-time control.
% In contrast, our work focuses on a unique combination of semantic spatial reasoning with policy learning in robotics, marking a distinct direction.

%% file: sec/3_method.tex
\section{Method}

\begin{figure*}[!h]
    \centering
    \includegraphics[width=0.85\textwidth]{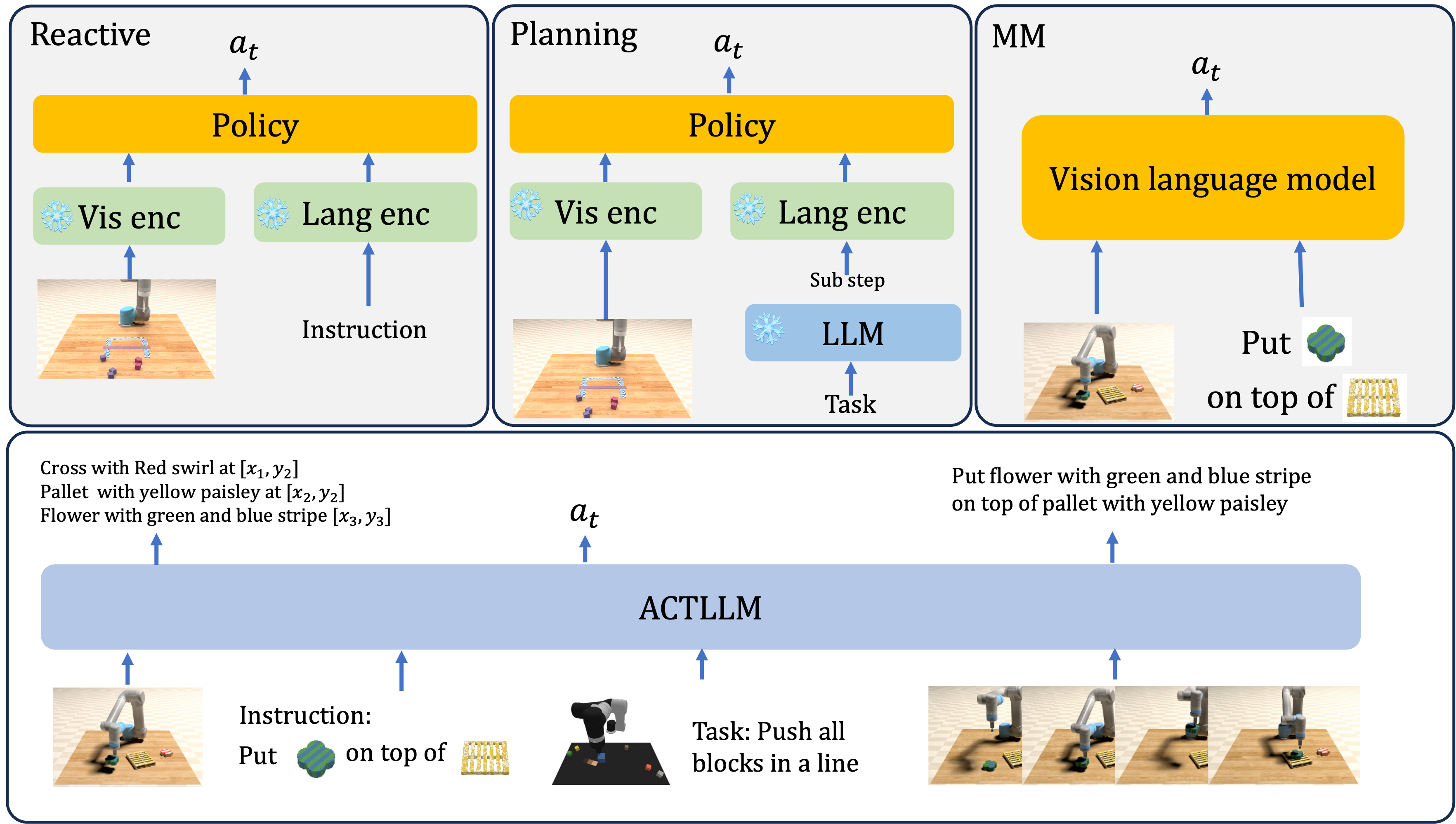}
    \caption{Comparison among our vision-language manipulation approach and existing solutions: \textbf{Reactive} indicates utilizing foundation models to extract features for following action generation~\cite{nair2022r3m}. In contrast, \textbf{Planning} strategies typically employ LLM to decompose tasks into sub-steps and solve them. \textbf{MM} means works directly fine-tuning vision-language models for robotic manipulation, as seen in~\cite{jiang2022vima, li2023mastering} directly fine-tune a vision-language model for robot manipulation. Our model can process both visual and language tokens as task-specific inputs, allowing us to generate actions that are more consistent and accurately correspond to the scene description.}
    \label{fig:compare}

\end{figure*}
\subsection{Preliminary}
We have access to a dataset $\mathcal{D} = \{\zeta_1, \zeta_2, \ldots, \zeta_n\}$ of $n$ expert demonstrations trajectory $\zeta$ with associated discrete-time input-action pairs $\zeta_i = \{(o_1, I_1, a_1), (o_2, I_2, a_2), \ldots\}$.
Notably, the instruction $I$ for each step may consist of text only, images only, or a combination of both as shown in Figure~\ref{fig:compare}
The action space consists of primitive motor skills, including actions such as \textit{pick and place}, \textit{wipe}, \textit{push}~\cite{zeng2021transporter}. 
Additionally, for each action, two positional vectors indicate the initial and target poses, denoted as $(p_{\text{initial}}, p_{\text{target}})$.

Our goal is to learn a policy model $\pi$ that can effectively generate an action $a_{t}$ to accomplish provided multimodal instruction $I$ based on the current execution history.
Inspired by recent advancements in imitation learning literacy~\cite{zare2023survey}, we designed our policy as $\hat{a} = \pi_\theta(x_t,x_{t+1})$. This all us to deduce the most likely action that transitions the agent from its current state to a subsequent state, where $x_t$ and $x_{t+1}$ represent the states extracted from observations.
For a one-step trajectory, represented as  $(o_t,a_t,o_{t+1}, I_t)$, we will map observations to their respective state representations through an observation model, yielding $x_t = \mathcal{M}(o_t), x_{t+1} = \mathcal{M}(o_{t+1})$.
During inference, to enable the policy leverage the future state $x_{t+1}$, we need to incorporate a forward model for forecasting future states given the current observation and instruction, formalized as $\tilde{x}_{t+1} = \mathcal{F_\beta}(o_t, I_t)$.
The essence of this formulation lies in the focus on the desired final state rather than the specific steps taken to achieve it, as multiple paths can lead to the same outcome.
In practical, this approach means that the model does not strictly penalize deviations from the ground-truth actions, as long as the alternative actions proposed lead to the same expected subsequent states.

\subsection{Structure scene description}
In optimizing our model, the key lies in the choice of state representation for both $\mathcal{M}$ and prediction $\mathcal{F}$ model. 
Our key insight for utilizing LLM as both a model $\mathcal{M}$ and a function $\mathcal{F}$, which not only enhances model optimization but also possesses a stable ground-truth, as previous approaches using hidden vectors for state representation can be unstable, particularly during the early training stages.

To enable LLM to provide structured representations rather than unstructured language, we leveraged predefined JSON schema to manipulate the decoding phase. 
During this phase, we pre-fill the fixed tokens of the data schema and delegate the generation of content tokens solely to the language model. 
This approach ensures that the structure remains consistent while the model dynamically fill the content.
Therefore, we define the state representation as a set of $\{(\mathfrak{o}_1, \mathfrak{c}_1, \mathbf{p}_1), (\mathfrak{o}_2, \mathfrak{c}_2, \mathbf{p}_2), (\mathfrak{o}_3, \mathfrak{c}_3, \mathbf{p}_3), \ldots\}$, where $\mathfrak{o}$, $\mathfrak{c}$ and $\mathbf{p}$ represent the object, color, and coordinates, respectively.

This design offers several benefits:
\begin{enumerate*}[label=(\roman*)]
    \item interpretability: this approach simplifies the understanding and validation of the model's decision-making process, making it more transparent how and why certain conclusions were reached. 
    \item clarity: we can formulate a clear loss function based on the state representation, rather than relying on ambiguous regression losses associated with hidden states.
    \item stable ground-truth: by leveraging access to the simulator's internal state, we can precisely obtain the condition of all objects.
\end{enumerate*}
To enable the LLMs to describe the scene with coordinates, we adopt the center position of the bounding box to denote the object's location.
Coordinates are normalized within a [0, 1] range and maintained to two decimal places; [0.50,0.50] indicates that the object is at the center of the view.
Since we would like to enlarge the token space of the LLM, we integrate the Adapter mechanism~\cite{houlsby2019parameterefficient} to model the spatial tokens. 
Once we have the scene decription, we can compute the loss as $ \mathcal{L}(x_t,x_{t+1},\tilde{x}_{t+1},\tilde{x}_{t})$

We adopt the approach from GLIP~\cite{Li_2022}, employing contrastive loss to compare predicted objects with ground truth text tokens. This involves directly matching predicted objects with their corresponding textual descriptors. The process includes:
\begin{enumerate*}[label=(\roman*)]
\item For each predicted object (query), we compute a dot product with the feature vectors of the text descriptions to produce logits corresponding to each text token, followed by applying focal loss to each logit.
\item Since the set of objects lacks inherent order, we use box regression and classification costs to perform bipartite matching between the predictions and the ground truth data.
\item Finally, we calculate the loss values for the matched predictions and their corresponding ground truths, ensuring precise alignment and accuracy.
\end{enumerate*}

By aligning both object properties and their locations with language tokens, we can enhance the model’s reliability and accuracy in scene prediction tasks.

\subsection{Action consistency loss}
We now focus on how to generate the action $a_t = \pi_\theta(x_t,x_{t+1})$ based on the state representation.
Concatenating another language model to predict actions based on generated or ground-truth state representations seems intuitive. However, it could introduce more drawbacks than benefits. This is primarily due to potential inaccuracies in scene descriptions during initial training stages, leading to noise in policy gradient updates.
Therefore, we propose constructing the policy by aggregating the sequence of token embeddings of the state representation using a learnable parameter $W$ with an attention module
\begin{align*}
Q = EW, \quad &x_{\text{agg}} = \text{softmax}(QQ^T)Q\\
\hat{a}_t &=  \text{MLP}(x_{\text{agg}})
\end{align*}
where \(E\) represents the sequence of token embeddings of $x_t,x_{t+1}$, \(E = [e_1, e_2, \ldots, e_n]\), and \(n\) is the number of tokens in the sequence.
Through this method, actions are derived from an aggregated token that adeptly encapsulates a sentence-level representation of the state.
Moreover, due to its lightweight design, the model exhibits better adaptability, which is crucial to smoothly accommodate improvements in state representations over time.

\begin{figure*}
	\centering
	\includegraphics[width=0.95\linewidth]{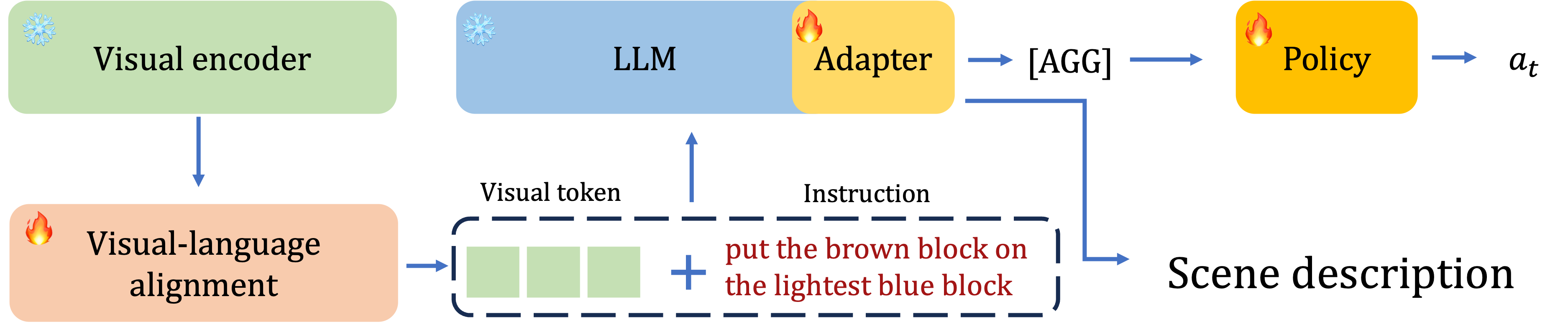}
	\caption{The illustration of our model. Our model consists of three main components: a visual encoder, a visual-language alignment layer, and a decoder-only LLM. The coordinates of a bounding box are converted into texts in a specific format. During training, we freeze the visual encoder and LLM and only update the Adapters and the alignment layer. Upon receiving an instruction, the LLM initially produces an aggregated token. This token informs the policy to generate the corresponding action for this step and also aids description of the environment.}

	\label{fig:relabel}
    \vspace{-3mm}
\end{figure*}

Once we have one step tuple $(s_t,a_t,s_{t+1}, I_t)$, we can design the optimized function to align the action with the state transition. The joint objective for training is formalized as:
\begin{equation}
\label{eq:loss}
\begin{aligned}
\min_{\theta,\beta} \; \mathcal{L}(x_t,x_{t+1},&\tilde{x}_{t+1},\tilde{x}_{t})\;+\; \mathcal{L}(a_t, \hat{a}_t) ,\\
\text{s.t.} \;\;\;\; 
&\tilde{x}_{t}, \tilde{x}_{t+1} = \mathcal{F}_\beta(o_t, I_t)\\
&\hat{a}_t = \pi_{\theta}(x_t, x_{t+1})
\end{aligned}
\end{equation}
where the action loss includes both the classification loss of primitive skill and the loss of the scene description.
As shown in Figure~\ref{fig:relabel}, the action consistency loss effectively harnesses additional information to more accurately align action and visual data together.
The advantage of this method of action generation is that it tightly integrates the action with the scene description, ensuring that the action is more closely associated with the context within the task and scene.
Moreover, by employing multi-turn tuning as detailed below, the action benefits from its correspondence to scene changes. Consequently, changes within the scene also influence how the action is generated.
\subsection{Markov decision making as visual dialogue}
Recent advancements, as highlighted in vicuna fintuning \cite{chiang2023vicuna}, demonstrate that multi-turn interactions significantly enhance Large Language Models (LLMs) by enabling more complex and context-rich dialogues.
Unlike single-turn dialogues, which generate responses without considering past interactions, multi-turn conversations draw upon previous exchanges, allowing the model to gain a comprehensive understanding of the context and the user's goals.

Expanding on the previously introduced concept, we propose a straightforward extension of our one-step optimization to a variable-length sequence of actions as shown in Figure~\ref{fig:conv}
This framework involves multi-turn dialogue tuning, where each step includes a current observation and instruction, leading to a unique state. The model needs to select the most suitable action based on the current state and overall task goal, incorporating the historical context of predicted states.

The joint loss is calculated at each time step and is optimized alongside the action prediction loss across the entire trajectory. 
The ﬁnal multi-step objective with feature space dynamics is defined as follows:
\begin{equation}
\label{eq:loss}
\begin{aligned}
\min_{\theta,\beta} \;\sum_{t=1}^{T-1}  \Big(\mathcal{L}(x_t,x_{t+1},&\tilde{x}_{t+1},\tilde{x}_{t})\;+\; \mathcal{L}(a_t, \hat{a}_t)\Big)  ,\\
\text{s.t.} \;\;\;\; 
&\tilde{x}_{t}, \tilde{x}_{t+1} = \mathcal{F}_\beta(o_t, I_t)\\
&\hat{a}_t = \pi_{\theta}(x_t, x_{t+1})
\end{aligned}
\end{equation}
This method allows the model to learn and adapt over time, fostering a more advanced and user-focused conversational experience. 
It enables users to switch tasks or offer prompt interventions in between, enhancing interaction quality and effectiveness.
As illustrated in Figure~\ref{fig:compare}, compared with other methods, our proposed model can process various types of information and generate essential output for robot manipulation.
One advantage of having a long context for the trajectory is that it ensures better alignment between the action and scene descriptions. Specifically, when tuning on the step tuple $(s_t,a_t,s_{t+1}, I_t)$, the action and scene descriptions are closely associated with the current state. However, when optimizing for multi-turn interactions, the action generation can leverage the extended context as cues. This allows for actions that consider a more comprehensive history.
\subsection{Implementation details}

Our model architecture is composed of two components
\begin{enumerate*}[label=(\roman*)]
    \item a MLLM serves as both $\mathcal{F}$ and $\mathcal{M}$. This component encompasses a visual encoder $\Phi_V$, a projection layer $\Phi_P$, and a Language Model $\Phi_L$. 
    \item a policy head $\pi$, which includes an attention module and an MLP.
\end{enumerate*}
The visual encoder is utilized to transform all images found in instructions or observations into a sequence of visual tokens, denoted as $Z_V = \Phi_V(I)$. 
To adapt this representation for use in the LLM, a linear layer $\Phi_P$ is employed, converting $Z_V$ into the LLM's input space, leading to $Z_T = \Phi_P(Z_V)$. 
Subsequently, both $Z_T$ and $Q_T$ are concatenated and passed through $\Phi_L$ to produce the scene description $x_t, x_{t+1}$. 
Finally, the policy head $\pi$ processes the scene description tokens and translates them into actions.
As discussed in \cite{wang2023makes}, fine-tuning the visual encoder, even with a modest visual instruction tuning dataset, can lead to semantic loss. This loss adversely affects the image representation capabilities of the visual encoder. Consequently, we opt to keep the visual encoder frozen, especially considering that simulations might exacerbate semantic loss.
We incorporate the LLAMA3-8B~\cite{dubey2024llama3herdmodels} model, as our LLM, specifically tuned for following instructions. 
To bridge visual tokens with the static LLM, we utilize an alignment layer. This is supplemented by an action projection module, which is crucial for converting model insights into actionable outputs.
AdamW is selected as the optimizer.
Further details will be provided in the supplementary material.

%% file: sec/4_exp.tex
\section{Experiment}
\label{sec:exp}

Our experiments focus on two benchmark
\begin{enumerate*}[label=(\roman*)]
    \item VIMA-BENCH: A newly introduced task suite and benchmark designed to facilitate the learning of general robot manipulation through multimodal prompts.
    \item CLIPORT: This dataset offers step-level text instructions paired with top-down RGB-D observations, serving as a platform for learning multi-task manipulation.
\end{enumerate*}
We carefully design evaluations to evaluate
\begin{enumerate*}[label=(\roman*)]
    \item Compositional Generalization which assesses the model's proficiency in handling objects with new shapes, colors, and entirely novel objects.
    \item In-context Generalization, where the model is evaluated with the novel tasks
\end{enumerate*}
Note that during the testing stage, we do not provide a state representation as input to the model. Therefore, we do not need to rely on any object detector to provide a structured screen description, as it is generated by the model.

\subsection{Compositional Generalization} 

Compositional generalization evaluates a model's ability to apply learned knowledge to new combinations of familiar elements or concepts. We demonstrate this in the both CLIPORT and VIMA-BENCH environment.

In the CLIPORT environment, we follow the PAFF framework \cite{ge2023policy}. We report the task success rate across 100 evaluation instances in 10 diverse scenes. These scenes feature various blocks, objects, and bowls, thereby testing the model's capacity to accurately place objects. We adopted two evaluation protocols as follows:
\begin{enumerate*}
    \item \texttt{pack-unseen-objects}: In this protocol, we train a policy to pack objects of different shapes into a brown box ('pack-shapes') and then evaluate its ability to 'pack-unseen-objects.'
    \item \texttt{put-shapes-in-bowls} Another evaluation involves placing blocks of different colors into bowls of different colors ('put-blocks-in-bowls') then we ask the model to put objects of different shapes into bowls of different colors.
\end{enumerate*}
We included all baseline models from PAFF, where MdetrORT~\cite{mdetr} is a variation of CLIPORT~\cite{cliport} by replacing the visual and language encoder, and AugORT~\cite{aug1} include more data augmentation to the MdetrORT.

\begin{table}[h]
% \begin{wraptable}{r}{0.5\textwidth} 
    \centering
    \caption{Results from the compositional and out-of-distribution generalization evaluations on the CLIPORT platform are presented. The primary metric for evaluation is the success rate. Each step is provided with a new instruction in the left column; a subsequent instruction is issued only after the previous one has been successfully executed.}

    \begin{tabular}{c|>{\centering\arraybackslash}p{1cm}|>{\centering\arraybackslash}p{1cm}|>{\centering\arraybackslash}p{1cm}|>{\centering\arraybackslash}p{1cm}}
        \toprule
        Method & \multicolumn{2}{c|}{put-shapes-in-bowls} & \multicolumn{2}{c}{pack-unseen-objects}\\
        \midrule
        CLIPORT & 28.0\% & 16.8\% & 58.9\% & 46.1\%\\
        MdetrORT & 33.8\% & 17.8\% & 62.0\% & 48.4\%\\
        AugORT & 34.4\% & 18.9\% & 63.1\% & 49.0\%\\
        PAFF & \textbf{51.0\%} & \textbf{35.0\%} & \textbf{72.8\%} & \textbf{63.8\%}\\
        ACTLLM & \textbf{64.0\%} & \textbf{66.2\%} & \textbf{85.8\%} & \textbf{79.6\%}\\
        \bottomrule
    \end{tabular}  
    \label{tab:cliport}
\end{table}

\begin{table*}[h]
    \caption {We conducted a comparative analysis of our methods using the VIMA-BENCH evaluation across four distinct levels. The ‘Avg’ represents the average success rate for all tasks within each level. To ascertain the success rate for each method, we sampled 200 episodes from each task. Our methods demonstrated significant improvements over baseline approaches.}\label{tab:main-result}
    \centering
    \scalebox{0.8}{
        \begin{tabular}{l|lllllllllllllllllllllllllll}
            \toprule
              & \multicolumn{6}{c}{L1} & \multicolumn{6}{c}{L2} & \multicolumn{6}{c}{L3} & \multicolumn{2}{c}{L4}    \\ 
            \cmidrule[0.5pt]{2-6} \cmidrule[0.5pt]{8-12} \cmidrule[0.5pt]{14-18}  \cmidrule[0.5pt]{20-21}
            Method  & Avg & T5 & T9 & T16 & T17 &
                    & Avg & T5 & T9 & T16 & T17 &
                    & Avg & T5 & T9 & T16 & T17 &
                    & Avg & T10 \\
            \midrule
             Gato         & 57.0 & 44.5 & 14.0 & 43.0 & 1.5 & 
                          & 53.9 & 46.0 & 10.5 & 42.0 & 1.0 & 
                          & 45.6 & 36.0 & 17.0 & 41.5 & 0.0 &
                          & 13.5 & 0.0 \\
             Flamingo     & 47.2 & 41.0 & 3.0  & 38.0 & 2.0 & 
                          & 47.1 & 43.0 & 4.5 & 40.0 & 1.0 & 
                          & 42.1 & 36.5 & 6.0 & 45.5 & 0.5 &
                          & 11.1 & 0.0 \\
             GPT          & 47.9 & 45.0 & 8.0  & 33.0 & 1.0 & 
                          & 47.4 & 43.0 & 10.5 & 34.0 & 3.0 & 
                          & 42.6 & 32.0 & 5.0 & 37.5 & 0.0 &
                          & 12.1 & 0.5 \\
            \midrule
             VIMA         & 87.2 & 65.0 & 13.5 & 88.0 & 77.0 & 
                          & 87.0 & 61.0 & 12.5 & 87.5 & 77.5 & 
                          & 84.0 & 63.0 & 12.0 & 58.5 & 78.0 &
                          & 49.6 & 0.0 \\

            \textbf{ACTLLM} & $\mathbf{90.5}$ & $\mathbf{78.3}$ & $\mathbf{65}$ & $\mathbf{96.0}$ & $\mathbf{86.0}$ & 
                          &  $\mathbf{90.9}$ & $\mathbf{78.2}$ & $\mathbf{61.5}$ & $\mathbf{93.0}$ &$\mathbf{84.3}$ & 
                          & $\mathbf{93.4}$ & $\mathbf{84.0}$ & $\mathbf{70.9}$ & $\mathbf{88.2}$ & $\mathbf{87.0}$ &
                          & $\mathbf{64.8}$ & $\mathbf{12.1}$ \\
    
             \bottomrule
        \end{tabular}
    }

    \vspace{-3mm}
    \label{table:vima}
\end{table*}

We present the evaluation results of compositional generalization in Table~\ref{tab:cliport}. 
Our method significantly outperforms the baseline across both evaluation protocols by a substantial margin, demonstrating the efficacy of our scene representation model in achieving superior compositional generalization.
Unlike PAFF, ACTLLM takes a comprehensive approach that goes beyond focusing solely on task-relevant objects; it considers all objects present. 
This broader perspective equips it to excel in tasks involving novel objects and shapes. 
Furthermore, ACTLLM can accurately generate scene descriptions containing various concepts, such as objects and containers, within a compositional setting. Additionally, it can learn object-aware representations without the need for specific objects and associated bounding boxes.

In VIMA-BENCH, the compositional generalization evaluation is more challenging can be broken down into three levels:
\begin{enumerate*}
  \item \textbf{Placement:} During training, all prompts are encountered, and only the placement of objects on the tabletop is randomized at testing.

  \item \textbf{Combinatorial:} In the training phase, all textures and objects are familiar, but during testing, new combinations of these elements are introduced.

  \item \textbf{Novel Object:} Both the test prompts and the simulated workspace feature novel textures and objects that were not encountered during training.
\end{enumerate*}
% In our study, we benchmark our approach against a diverse set of baseline methods as introduced in the original VIMA paper.
We include results for  models such as Gato~\cite{reed2022generalist}, Flamingo ~\cite{alayrac2022flamingo}, and GPT~\cite{jiang2022vima} as reported directly within the VIMA paper.
% Gato: This model is a decoder-only architecture designed to handle tasks across various domains. It operates by processing sequences of observations and actions through prompting.
% Flamingo: As a vision-language model, Flamingo produces textual outputs in reaction to multimodal prompts. It leverages the Perceiver architecture for encoding an arbitrary number of images into a fixed token representation, which then informs the language decoding process via cross-attention mechanisms.
% GPT: Employing a decoder-only approach, GPT processes tokenized multimodal prompts to autoregressively generate subsequent actions, guided by provided instructions and interaction histories.
% Similar to our method, these models engage directly with raw image observations. Additionally, we explore results from object-centric approaches, including VIMA, which differ by utilizing ground truth object crops rather than raw frames for observation.
% VIMA utilizes a pre-trained T5 model to encode multimodal prompts, subsequently directing a robot controller via cross-attention layers.
The results of the evaluation for various levels of evaluation protocols are presented in Table~\ref{table:vima}. Since we have limited space, we only report the success rates of representative tasks for which different methods show significant performance differences. The average (Avg) indicates the success rate of all tasks at a particular evaluation level. We can see that our methods outperform all baseline methods in the first three levels, which require the learning of new combinations of familiar elements or concepts.

\subsection{In-context Generalization}

% Beyond mastering compositional generalization mainly designed for objects, another critical aspect in robotic manipulation is the ability to learn and perform new tasks.
This challenge is newly introduced in VIMA,  where new tasks are defined by novel prompt templates during test phases.  
These templates encompass not only novel actions but also novel objects that were not present in the training data.
Our evaluation follows a two-fold approach: first, we follow the VIMA protocol, which involves directly executing Level 4 tasks, as shown in Table~\ref{table:vima}; second, we use in-context demonstration 
videos that convey the essence of the tasks during testing. 
The inclusion of in-context learning, specifically during testing phases, is aimed at examining the model's zero-shot adaptation capabilities. 
This approach underscores the importance of providing models with dynamic, real-world contexts that enhance their learning and adaptation processes.
Following VIMA, we save T9: Twist T10: Follow Motion T11: Follow Order as a novel task. 
We then conducted an ablation study to assess the effectiveness of various modules. 

  % The table will be placed on the right side of the text.
\begin{table}[h]
\caption{Evaluating the in-context generalization capability of the ACTLLM with \emph{Twist} and \emph{Follow Order} as novel testing tasks.}
\centering
\scalebox{0.9}{
    \begin{tabular}{l|ccc|c}
        \toprule
        Task &  T9   & T10  & T11 & Overall \\
        \midrule
        Our Method & $\mathbf{20.1\%}$ & $\mathbf{63.4\%}$ & $\mathbf{10.0\%}$     & $\mathbf{36.2 \%}$ \\
        \midrule
        w/o task tuning & 14.5\%  & 38.9 \% & 6\%  & 26.7\% \\

        w/o future state & 12.4\%  & 23.9 \% & 4.7\%  & 22.3\% \\

        w/o multi-turn & 16.1\%  & 29.9 \% & 3.5\%  & 24.3\% \\
        \bottomrule
    \end{tabular}
}

\label{tab:icl-results}
\end{table}
% \begin{table}[h]
%     \caption{Evaluating the in-context generalization capability of the ACTLLM with \emph{Twist} and \emph{Follow Order} as novel testing tasks.}
%     \centering
%     \scalebox{0.9}{
%         \begin{tabular}{l|ccc|c}
%             \toprule
%             Task &  T9   & T10  & T11 & Overall \\
%             \midrule
%             Our Method & $\mathbf{20.1\%}$ & $\mathbf{63.4\%}$ & $\mathbf{10.0}$ \%     & $\mathbf{36.2 \%}$ \\
%             \midrule
%             w/o task tuning & 14.5\%  & 38.9 \% & 6\%  & 26.7\% \\

%             w/o future state & 12.4\%  & 23.9 \% & 4.7\%  & 22.3\% \\

%             w/o multi-turn & 16.1\%  & 29.9 \% & 3.5\%  & 24.3\% \\
%             \bottomrule
%         \end{tabular}
%     }
%     \label{tab:icl-results}    
% \end{table}
As shown in table~\ref{tab:icl-results}, removing task tuning significantly reduces performance across all tasks, leading to a marked decrease in the accuracy of in-context learning. This suggests the model struggles to interpret tasks from videos, a foreseeable outcome since the training primarily incorporates multimodal prompts rather than video content. Such findings underscore the critical role of task tuning in enhancing the model's ability to generalize.
The omission of the future state prediction module severely impacts performance, particularly in task T10, underscoring the importance of the model’s predictive capabilities for effective action planning. Additionally, removing multi-turn tuning also reduces effectiveness, though less so than the absence of future state prediction, which likely serves as a regularization mechanism. 

%% file: sec/5_conclusion.tex
\section{Conclusion, future work and limitation}
% This paper explores fine-tuning MLLM  for improved robotic manipulation guided by multimodal instructions. Our approach, ACTLLM, using the Vicuna Large Language Model, achieves top-notch results on a benchmark dataset. Our experiments highlight the benefits of incorporating an action consistency loss, especially in enhancing Compositional Generalization and zero-shot generalization.

% Our work contributes to the goal of creating intelligent robotic systems that can understand and respond to human language commands intuitively, enhancing human-robot interaction. 
% However, a current limitation, as demonstrated in the experimental section, is that the model still struggles to accurately recognize the shapes of objects, demanding greater precision in the grounding ability of MLLMs.
% Looking forward, there is an exciting opportunity for future research to explore how our model performs in real-world settings and its in-context learning capabilities, potentially expanding the use of intelligent robotic systems across various domains.

Our method, ACTLLM, demonstrates superior performance on a benchmark dataset by incorporating an action consistency loss, notably improving compositional and zero-shot generalization. Our research advances the development of intelligent robotic systems that intuitively understand and execute human language commands, improving human-robot interaction. Future research could explore real-world applications and in-context learning capabilities, broadening the deployment of intelligent robotic systems.

% \section{Limitation}
While our model shows promising potential, a key challenge remains: 2D images often fail to capture the precise 3D locations and desired states essential for effective robot control.

% Additionally, when using the VIMA dataset, objects depicted without color reference in the images default to gray in appearance. This scenario is atypical in real-world testing environments, where such representations are seldom encountered.

% Improvements in these areas are essential for enhancing performance and extending the practical application of our model in complex, real-world scenarios.